\documentclass[
]{ceurart}

\usepackage{listings}

\usepackage{subcaption} 
\usepackage{graphicx}
\usepackage{algorithmic} 
\usepackage{algorithm}

\usepackage{amsthm} 

    \theoremstyle{plain}

    \theoremstyle{definition}

    \theoremstyle{note}

\begin{document}

\copyrightyear{2024}
\copyrightclause{Copyright for this paper by its authors.
Use permitted under Creative Commons License Attribution 4.0 International (CC BY 4.0).}

\conference{AICS'24: 32nd Irish Conference on Artificial Intelligence and Cognitive Science, December 09--10, 2024, Dublin, Ireland}

\title{Exploring the Potential of Bilevel Optimization for Calibrating Neural Networks}


\author[1]{Gabriele Sanguin}[%
orcid=,
email=gabriele.sanguin@math.unipd.it,
]
    \address[1]{Dipartimento di Matematica, Universit\`a degli Studi di Padova, Via 8 Febbraio 2, 35122 Padova (Italy)}

\author[2]{Arjun Pakrashi}[%
orcid=0000-0002-9605-6839,
email=arjun.pakrashi@ucd.ie,
]
    \address[2]{School of Computer Science, University College Dublin, Belfield, Dublin 4 (Ireland)}

\author[3]{Marco Viola}[%
orcid=0000-0002-2140-8094,
email=marco.viola@dcu.ie,
]
\cormark[1]
\address[3]{School of Mathematical Sciences, Dublin City University, Collins Avenue Ext, Dublin 9 (Ireland)}

\author[1]{Francesco Rinaldi}[%
orcid=0000-0001-8978-6027,
email=rinaldi@math.unipd.it,
]

\cortext[1]{Corresponding author.}

\begin{abstract}
Handling uncertainty is critical for ensuring reliable decision-making in intelligent systems. Modern neural networks are known to be poorly calibrated, resulting in predicted confidence scores that are difficult to use. This article explores improving confidence estimation and calibration through the application of bilevel optimization, a framework designed to solve hierarchical problems with interdependent optimization levels. A self-calibrating bilevel neural-network training approach is introduced to improve a model's predicted confidence scores. The effectiveness of the proposed framework is analyzed using toy datasets, such as Blobs and Spirals, as well as more practical simulated datasets, such as Blood Alcohol Concentration (BAC). It is compared with a well-known and widely used calibration strategy, isotonic regression. The reported experimental results reveal that the proposed bilevel optimization approach reduces the calibration error while preserving accuracy.\end{abstract}


\maketitle
\vspace{-1.1em}
\section{Introduction}
Machine learning classification algorithms with increasingly higher discriminative power, especially neural networks, have rapidly developed in the last decade. Such advanced models are generally supposed to help humans make decisions by assisting. Although these models have a high discriminative power, sometimes they will predict something completely incorrect with a fairly high confidence score. This creates a huge problem in highly regulated and sensitive real-world applications (e.g. medical field, autonomous vehicles, healthcare diagnostics, financial forecasting, etc.) \cite{Minderer2021neurips}. Therefore, it is important for these models to provide a meaningful confidence, based on which it can say "I don't know", when they are not confident enough, so that the human expert can inspect, and make further decisions. This is sometimes called \emph{learning to reject} or \emph{abstention} \cite{zhang2023survey,hendrickx2024machine}.

\emph{Confidence} is a probabilistic score that a model assigns to each prediction of a data point, and it determines how certain the model is about the prediction. One straightforward approach to use such a confidence score to understand if the model is confident \emph{enough} or not is to define an interval, a \emph{rejection window}, within which if the prediction falls, it will be marked as \emph{reject}. It is however hard to fix such a confidence window to decide the rejection window for such models, especially modern neural network models, because such models are known to have poor model confidence calibration \cite{Minderer2021neurips,guo2017calibration}.

Confidence \emph{calibration} is the process of adjusting the confidence scores to better align with the actual likelihood of correctness. 
Well-calibrated confidence is crucial for effective decision making and is important for the interpretation of the model, since humans have a natural cognitive intuition for probabilities \cite{cosmides1996humans}. Accurate confidence scores make it easier for users to comprehend how confident the model was during prediction and to establish trust on the decisions being made. Moreover, accurate confidence scores are essential if such a rejection window needs to be defined by a human expert.

There are two types of confidence calibration. Firstly, \emph{post-calibration}, where the output scores/probabilities of the main model are re-adjusted by another external calibration model. For such methods, the main model does not need to be modified. In \cite{guo2017calibration}, the authors have demonstrated how well-known post-calibration methods can be used to calibrate existing models. On the other hand, \emph{self-calibration} method are algorithms which integrate the confidence calibration process into the model training itself. These methods aim to ensure that the model's probabilities are calibrated during the training phase, without the need for a separate calibration step. 

The objective of this article is to present an initial study on whether it is possible to self-calibrate neural network models by exploiting \emph{bilevel optimization} (BO), a mathematical framework specifically designed to solve hierarchical two-level decision-making optimization problems. BO has recently gained importance in machine learning, particularly in hyperparameter optimization and metalearning \cite{pedregosa2016hyperparameter,franceschi2017forward,franceschi2018bilevel}. The choice of such a framework is obvious where the inner-level optimization problem tackles the training of the model, while the outer-level optimization problem addresses the model confidence calibration.


To the best of our knowledge, in the context of uncertainty scores, the only work in the literature that uses BO is \cite{jain2022selective}. Here, we define a BO framework to train two different architectures, one for classification and one for the uncertainty score (which is then tested as a rejection function), thus leading to a significant increase of the parameters to be trained. Then we focus on the study on the selective potential of such a score.
The work presented in the current article is the first of its kind, which is quite distinct from the one in \cite{jain2022selective}. The objective of the current work is to propose a BO framework to train a single self-calibrated deep neural network, \emph{BO4SC}, and provide an initial, but crucial analysis of the applicability of the approach.

The article is organized as follows. Section \ref{sec:theory} discusses the mathematical foundations of confidence estimation, calibration methods, and bilevel optimization (BO). Section \ref{sec:BO4SC} introduces BO4SC, a BO framework for confidence estimation in neural networks. In Section \ref{sec:experiments} the initial experiments and analysis are shown and discussed. Finally, Section \ref{sec:conclusions} concludes the article.

\section{Theoretical Background \label{sec:theory}}
This section will briefly introduce and discuss the relevant parts of confidence estimation, model calibration evaluation, calibration methods, and bilevel optimization.

\subsection{Confidence Estimation}
\label{confidence_estimation}
For a machine learning model, the standard process (referred to as \emph{standard} method later in Section~\ref{sec:experiments}) to accurately predict an output involves learning a mapping function that can generalize well from training data to unseen samples. Let \(\{X, Y\} = \{(\mathbf{x}_i, y_i)\}_{i=1}^n\) be a labeled dataset, where \(\mathbf{x}_i\) represents a data point and \(y_i \in \{1, 2, \dots, C\}\) is its corresponding class label, where \(C\) being the total number of classes and $n$ the number of data points. The objective is usually to learn a function \(f\) that maps an unseen data point \(\mathbf{x}_t\) to a predicted output \(\hat{y}_t = f(\mathbf{x}_t)\). This function \(f\) can be efficiently and effectively trained by minimizing the empirical loss over all training data.


Confidence estimation involves assigning a probabilistic score to each prediction which reflects the model's certainty about the predicted output. \emph{Confidence} is a score function $\hat{p_i} = g(f,\mathbf{x}_i, \hat{y_i}),$
which measures the likelihood of the prediction $\hat{y}_i$ being correct given the features $\mathbf{x}_i$ and the classifier~$f$. Ideally, the confidence score should be continuous and fall within the range $[0, 1]$.

A variety of methods have been developed to enable confidence estimation across different model types. Among these approaches, we find
\textit{distance-based} methods, which use the distance of a data point from other points, decision boundaries, or centroids of classes to estimate confidence \cite{weinberger2009distance,mendes2017nearest,jiang2018trust,mandelbaum2017distance,papernot2018deep}.
\textit{Bayesian uncertainty} methods use Bayesian principles to model uncertainty, providing a probabilistic interpretation of confidence \cite{gal2016dropout,blundell2015weight,kristiadi2020being,riquelme2018deep}.
\textit{Reconstruction error} techniques rely on the error of reconstructing input data to obtain a confidence score, often used in models with an encoder-decoder framework, such as autoencoders. The idea is that a high reconstruction error indicates a lower confidence in the prediction of the model, \cite{xia2015learning,yoshihashi2019classification}.
\textit{Ensemble methods}  utilize the variance among predictions from multiple models to estimate confidence \cite{srivastava2014dropout,lakshminarayanan2017simple}.
\textit{Extreme value theory (EVT)} approaches are based on EVT that assess confidence by modeling the tail distributions of prediction scores \cite{bendale2016towards}.
Finally, \textit{logits-based} techniques involve the use of logits, which include probabilistic outputs and other mechanisms derived from the raw scores produced especially by neural networks. These scores can be transformed or analyzed to estimate the confidence of the predictions \cite{de2000reject}.

Logits-based methods are the ones that are gaining more attention for the recent extensive use of neural networks. The experiments in the current work use a smooth version of \emph{maximum class probability} (MCP), which is a common approach in many classification tasks.
We define the MCP as follows
\begin{equation}
    \hat{p}(x) = \max_{c} P(y = c \mid \mathbf{x}),
\end{equation}
where \(P(y = c \mid \mathbf{x})\) represents the predicted probability of class \(c\) for input \(\mathbf{x}\) after applying a softmax function to the logits.

\subsection{Evaluating Model Calibration}
\label{calibration_metrics}
Unlike classification functions, which can be efficiently learned from labeled data, there is no available supervisory information for directly learning a \emph{confidence function}
and the first challenge is how to estimate it from pre-trained models (e.g. with MCP). Unfortunately, in many cases, especially in modern deep neural networks, the calculated confidences tend to be overestimated, meaning that the models are \emph{over-confident} (see \cite{zhang2023survey}). This can be formally described as
\begin{equation}
\label{eq:overconfident}
    P(\hat{y} = y \mid \hat{p} = p) < p,
\end{equation}
where $p$ represents the true probability.

To address this issue, it is necessary to employ methods to calibrate the confidence. A model is considered \emph{calibrated} if \(\hat{p}_i\) accurately reflects the true likelihood of correctness:
\begin{equation}
\label{eq:calibrated}
    P(\hat{y} = y \mid \hat{p} = p) = p, \quad \forall p \in [0, 1].
\end{equation}

To understand whether a model is well-calibrated one can exploit metrics quantifying the degree to which the model's predicted probabilities align with the actual outcomes or its likelihood. It is important to note that there is no single, universally accepted metric for assessing calibration. 

In this work, we will use two of the most common calibration metrics in the literature, namely: \textit{reliability diagrams} and \textit{expected calibration error} (ECE). 

\emph{Reliability diagrams} are a visual tool used to assess model calibration \cite{degroot1983comparison,niculescu2005predicting} by plotting the expected accuracy of samples against their predicted confidence levels. The predictions are grouped into \(M\) interval bins, each of size \(\frac{1}{M}\). By letting \(B_m\) represent the set of indices of samples whose predicted confidence falls within the interval \(I_m = \left(\frac{m-1}{M}, \frac{m}{M}\right]\), the accuracy for bin \(B_m\) is calculated as
\[
\text{acc}(B_m) = \textstyle \frac{1}{|B_m|} \sum_{i \in B_m} \mathbf{1}(\hat{y}_i = y_i),
\]
where \(\hat{y}_i\) and \(y_i\) are the predicted and true class labels for data point \(\mathbf{x}_i\), respectively. According to basic probability theory, \(\text{acc}(B_m)\) serves as an unbiased and consistent estimator of \(P(\hat{y}~=~y~\mid~\hat{p}~\in~I_m)\).

The average confidence within the bin \(B_m\) is given by
\[
\text{conf}(B_m) = \textstyle \frac{1}{|B_m|} \sum_{i \in B_m} \hat{p}_i,
\]
where \(\hat{p}_i\) represents the confidence of the sample \(i\). For a perfectly calibrated model, the relationship \(\text{acc}(B_m) = \text{conf}(B_m)\) should hold for all \(m \in \{1, \ldots, M\}\), i.e., the plot should follow the identity line. It is important to note that reliability diagrams do not display the proportion of samples in each bin. This is also why they are often paired with a density plot of confidence prediction, called \emph{confidence histograms}.

The ECE represents the weighted average of the absolute difference between accuracy and confidence over all prediction bins. Formally, it is defined as:
\begin{equation}
\text{ECE} = \textstyle \sum_{m=1}^{M} \frac{|B_m|}{n} \left| \text{acc}(B_m) - \text{conf}(B_m) \right|,
\end{equation}
where \( n \) is the total number of samples. Although ECE is widely adopted due to its simplicity and interpretability, it is sensitive to the choice of the number of bins \( M \), which can affect the accuracy of the measurement.

\subsection{Calibration Methods}
\label{calibration_methods}
Calibration methods can be categorized into two types, namely post-calibration and self-calibration.

\textbf{Post-calibration}: These methods involve adjusting the output probabilities of a pretrained model using a separate calibration model, applied after the initial model has been trained. This adjustment aims to align the predicted probabilities with the true likelihood of events.

Among the most common approaches we can find histogram binning \cite{zadrozny2001obtaining}, Bayesian binning into quantiles (BBQ) \cite{naeini2015obtaining}, Platt scaling \cite{platt1999probabilistic,niculescu2005predicting} and its derivative matrix and vector scaling and temperature scaling \cite{guo2017calibration}.
Other recent methods include beta calibration \cite{kull2017beta}, Shape-Restricted Polynomial Regression \cite{wang2019calibrating} and neural calibration \cite{pan2020field}.

The experiments in the current work make use of the \emph{isotonic regression} \cite{zadrozny2002transforming}, because of its simplicity and effectiveness. Isotonic regression learns a piecewise constant function \( f \) to transform uncalibrated outputs into calibrated ones, by minimizing the squared loss subject to the constraint that \( f \) is a non-decreasing function.


\textbf{Self-calibration}: These methods integrate the calibration process into the model training itself. These methods aim to ensure that the model's probabilities are calibrated during the training phase, without the need for a separate calibration step. Self-calibration often requires modifying the loss function or the training procedure to directly incorporate the calibration objectives.
Techniques such as Bayesian neural networks, which incorporate uncertainty directly into the model predictions through probabilistic inference, inherently produce better-calibrated probabilities \cite{blundell2015weight,kwon2020uncertainty}.

The main objective of this article is to explore a new self-calibration strategy for neural networks that makes use of a bilevel optimization framework.
\subsection{Bilevel Optimization}

Bilevel optimization (BO) is a mathematical approach designed to address hierarchical decision-making processes, where decisions made at an outer level influence the outcomes of an inner level, which in turn affects the outer level. This hierarchical structure is prevalent in many real-world scenarios, such as economics, engineering, management, and various public and private sector operations.
The distinctive feature of bilevel optimization lies in its two interconnected levels of optimization. Each level has its own objectives and constraints, and there are two classes of decision vectors: the leader's (outer-level) decision vectors and the follower's (inner-level) decision vectors. The inner-level optimization is a parametric optimization problem solved with respect to the inner-level decision vectors, while the outer-level decision vectors act as parameters. The inner-level optimization problem acts as a constraint to the outer-level optimization problem, such that only those solutions are considered feasible that are optimal for the inner level.





By denoting the outer and inner parameters as \( \mathbf{w} \) and \( \theta \), respectively, we can define an unconstrained BO problem as
\begin{equation}
\min_{\mathbf{w}} \;\; f(\mathbf{w}, \theta^*) \qquad \text{s.t.} \;\; \theta^* \in \arg \min_{\theta} g(\mathbf{w}, \theta),
\label{eq:bo_problem}
\end{equation}
where \( \theta^* \) is one of the minimizers of \( g \).


\emph{Gradient-based} approaches are now the most commonly used methods for solving bilevel optimization problems. The most compelling approach to gradient-based bilevel optimization is to replace the inner problem with a \emph{dynamical system}. This idea, discussed, e.g., in \cite{franceschi2017forward,domke2012generic,maclaurin2015gradient}, involves approximating the bilevel problem with a sequence of optimization steps, which allows for efficient gradient computation.

Specifically, consider a prescribed positive integer $T$ and let $[T] = \{1, 2, \dots, T\}$. We now rewrite the bilevel problem Eq.\eqref{eq:bo_problem} with the following approximation:

\begin{equation}
\begin{aligned}
\min_{\mathbf{w}} \;\;& f(\mathbf{w}, \theta^T(\mathbf{w})) \\
\text{s.t. } & \theta^0 (\mathbf{w}) = \Phi_0(\mathbf{w}),\\
& \theta^t (\mathbf{w}) = \Phi_t (\theta^{t-1} (\mathbf{w}), \mathbf{w}), \quad t \in [T],
\end{aligned}
\label{eq:bilevel_problem_2}
\end{equation}

\noindent where $ \Phi_0 : \mathbb{R}^n \rightarrow \mathbb{R}^m$ is a smooth initialization mapping and for each \( t \in [T] \) , \( \Phi_t: \mathbb{R}^m \times \mathbb{R}^n \rightarrow \mathbb{R}^m \) represents the operation performed by the \( t \)-th step of an optimization algorithm. For example, if the optimization dynamics is gradient descent, we might have:

\begin{equation}
\Phi_t(\theta^{t-1}, \mathbf{w}) = \theta^{t-1} - \eta_t \nabla_{\theta} g (\theta^{t-1}, \mathbf{w}),
\end{equation}
\noindent where \( (\eta_t)_{t \in [T]} \) is a sequence of step sizes.

This approach approximates the bilevel problem and gives the possibility to use gradient descent also to solve the outer objective. To this end, one has to compute an \emph{hypergradient}, which is the gradient of the outer objective \( f( \mathbf{w}, \theta^T (\mathbf{w})) \) with respect to the hyperparameters \( \mathbf{w} \), i.e.,

\begin{equation}
    \nabla_{\mathbf{w}} f( \mathbf{w}, \theta^T (\mathbf{w})) = \nabla_{\mathbf{w}} f( \mathbf{w}, \theta^T) + 
    [J_{\theta^T(\mathbf{w})}(\mathbf{w})]^\top \nabla_{\theta} f( \mathbf{w}, \theta^T),
\end{equation}
where rows in the Jacobian matrix $J_{\theta^T(\mathbf{w})}(\mathbf{w})$ contain gradients of the entries of $\theta^T$ with respect to $\mathbf{w}$. 

The reformulation \eqref{eq:bilevel_problem_2} allows for efficient computation of the hypergradient using reverse or forward mode algorithmic differentiation.

\section{BO4SC: A Bilevel Optimization Framework for Self-Calibration\label{sec:BO4SC}}

We introduce here the bilevel optimization framework we designed to enhance confidence estimation, which we will name \textit{BO4SC}.

We here assume that the prediction models are characterized by a \emph{dual-output structure}: one output to provide the prediction for the data point, the other to estimate the confidence of that prediction. This is essential because we want both the class predictions and the confidence estimation to be dependent on the same model parameters.
For the model $m$, parametrized by $\theta$, we will denote the output relative to the sample $\mathbf{x}_i$ with

\begin{equation}
    m(\mathbf{x}_i, \theta) = ( \hat{y}(\mathbf{x}_i, \theta), \hat{p}(\mathbf{x}_i, \theta)) = (\hat{y}_i, \hat{p}_i),
\end{equation}

where $\hat{y}_i$ is the class prediction and $\hat{p}_i$ is his confidence estimation.

Now consider the optimization problem in Eq. \eqref{eq:bo_problem}, with the outer parameters the weights $\mathbf{w}$ and the inner parameters $\theta$ of the model $m_\theta$. The inner loss function $g$ is trained on the training set (\(D^{\text{train}}\)), focusing on minimizing the \emph{weighted} \textit{cross-entropy} (CE) loss over the model's prediction output with the objective of minimizing the model's parameters $\theta$:

\begin{equation}
     g(\mathbf{w}, \theta) = \frac{1}{|D^{\text{train}}|} \sum_{i \in D_{\text{train}}} w_i \cdot \text{CE}(\hat{y}(\mathbf{x}_i, \theta), y_i)
\end{equation}

supposing \(\theta^*\) to be unique and where the CE loss is defined as:

\begin{equation}
    \text{CE}(\hat{y}(\mathbf{x}_i, \theta), y_i) = \textstyle -\sum_{c=1}^{C} y_{i,c} \log(\hat{y}(\mathbf{x}_i, \theta)_c)
\end{equation}

Here, \(C\) represents the number of classes, \(y_{i,c}\) is the binary indicator (0 or 1) if the class label \(c\) is the correct classification for input \(\mathbf{x}_i\), and \(\hat{y}(\mathbf{x}_i, \theta)_c\) is the final logit for class \(c\) given input \(\mathbf{x}_i\) according to the model \(\hat{y}(\cdot, \theta)\).

The outer loss function \(f\), on the other hand, is evaluated on the validation set (\(D^{\text{val}}\)), where it aims to minimize a \textit{binary cross-entropy} (BCE) loss on the model's confidence output $\hat{p}(\cdot, \theta)$. The objective is to learn weights for each sample in the training set that can effectively balance the trade-off between prediction accuracy and confidence calibration:

\begin{equation}
 f(\mathbf{w}, \theta^*) = \frac{1}{|D^{\text{val}}|} \sum_{j \in D^{\text{val}}} \text{BCE}(\hat{p}(\mathbf{x}_j, \theta^*_{\mathbf{w}}), y_j),
\end{equation}
\noindent where \(\theta^*_{\mathbf{w}}\) are the model parameters found by the inner problem and that depend on the weights \(\mathbf{w}\) assigned to the training samples. \(\hat{p}(\cdot, \theta)\) is the confidence output of the model.

The binary cross-entropy (BCE) loss is defined as:

\begin{equation}
    \text{BCE}(\hat{p}(\mathbf{x}_j, \theta^*), y_j) = - \left[ y^B_j \log(\hat{p}(\mathbf{x}_j, \theta^*)) + (1 - y^B_j) \log(1 - \hat{p}(\mathbf{x}_j, \theta^*)) \right]
\end{equation}

In this equation, \(y^B_j\) is the true \emph{binary label} ($0$ or $1$) for the sample \(\mathbf{x}_j\), indicating whether \(\mathbf{x}_j\) has been correctly classified (i.e. $\hat{y}_j = y_j$); \(\hat{p}(\mathbf{x}_j, \theta^*)\) represents the predicted confidence (probability) that \(\mathbf{x}_j\) belongs to the positive class according to the model.

The difficulty in solving this bilevel optimization problem usually lies in the accurate computation of the hypergradient \(\nabla_{\mathbf{w}} \mathcal{L}_{\text{outer}}(\mathbf{w}) = \nabla_{\mathbf{w}} f(\mathbf{w}, \theta^*_\mathbf{w})\), which necessitates sophisticated approaches requiring a large cost in time and memory performance.

We schematize as Algorithm~\ref{alg:hypergradient_descent_for_ltr} the approximate hypergradient descent algorithm we implemented to solve the BO4SC problem.

\begin{algorithm}[htp]
\caption{BO4SC via Approximate Hypergradient Descent}
\label{alg:hypergradient_descent_for_ltr}
\begin{algorithmic}

\STATE \textbf{Initialize:} Set initial weights \ensuremath{\mathbf{w}^0} and model parameters \ensuremath{\theta^0}.

\FOR{$j = 0,1,\ldots$}

\FOR{$k = 0$ to $T-1$}
\STATE \COMMENT{Inner loop: gradient descent on inner loss}

\STATE Compute the gradient of the inner loss w.r.t. \(\theta^k\):

\begin{equation*}
\nabla_{\theta} g(\mathbf{w}^j, \theta^k) = \frac{1}{|D^{\text{train}}|} \sum_{i \in D^{\text{train}}} w_i^j \cdot \nabla_{\theta} \text{CE}(\hat{y}(\mathbf{x}_i, \theta^k), y_i)
\end{equation*}

\STATE Update model parameters \(\theta^k\) using gradient descent:
$\quad\theta^{k+1} = \theta^k - \eta_{\theta} \cdot \nabla_{\theta} g(\mathbf{w}^j, \theta^k)$

\ENDFOR

\STATE Set \(\theta^j_{\mathbf{w}} = \theta^T\) \COMMENT{Final inner solution after $T$ iterations, in function on outer parameters $\mathbf{w}$}

\STATE Compute the \emph{hypergradient}, i.e. the gradient of the outer loss w.r.t. \(\mathbf{w}\), using the approximated \(\theta^j_{\mathbf{w}}\):

\begin{equation}
\label{eq:outer_loss_gradient}
\nabla_{\mathbf{w}} f(\mathbf{w}, \theta^j_{\mathbf{w}}) = \frac{1}{|D^{\text{val}}|} \sum_{j \in D^{\text{val}}} \nabla_{\mathbf{w}} \text{BCE}(\hat{p}(\mathbf{x}_j, \theta^j_{\mathbf{w}}), y_j)
\end{equation}

\STATE Update the outer-parameters \(\mathbf{w}^j\) using gradient descent:
$\quad \mathbf{w}^{j+1} = \mathbf{w}^j - \eta_{\mathbf{w}} \cdot \nabla_{\mathbf{w}} f(\mathbf{w}^j, \theta^j_{\mathbf{w}})$
\ENDFOR

\end{algorithmic}
\end{algorithm}


\section{Experiments and Results \label{sec:experiments}}
In this section we present our experiment process and the results. First, we give an overview on the training approaches we compared and the datasets we used. Then we analyse the experiment results.

This work mainly focuses on the proposed method along with two others which are described as follows:

\begin{itemize}
    \item \emph{Standard}: this refers to the standard training procedure in which the model's parameters are updated using backpropagation based on a \emph{single} loss function.

\item \emph{Isotonic Regression (IsoReg)}: it is the non-parametric method used to calibrate confidence scores after the initial training phase of a model with the \emph{Standard} method (see Section \ref{calibration_methods}).

    \item \emph{BO4SC}: the proposed method in this work Algorithm~\ref{alg:hypergradient_descent_for_ltr}).
\end{itemize}

In the implementation of the BO4SC algorithm, particularly for the explicit calculation of the outer loss gradient with respect to \(\mathbf{w}\) (that is, the gradient of \(\theta^T_{\mathbf{w}}\) with respect to \(\mathbf{w}\)), the Python package \texttt{torchopt} \cite{JMLR:TorchOpt} was used. \texttt{torchopt} is a library that extends PyTorch \cite{paszke2017automatic} by providing tools for higher-order optimization, specifically tailored for problems involving complex optimization hierarchies such as bilevel optimization. It enables efficient computation of the hypergradients. By leveraging \texttt{torchopt}, we can accurately and efficiently compute the required gradients in Eq. \eqref{eq:outer_loss_gradient}, thereby facilitating the optimization process in our experiments.

To facilitate the initial investigation, some toy datasets were built which were used as a diagnostic tool to understand BO4SC behavior, as well as how it compares with others. These datasets have two features to facilitate visual inspection.

%
%

The first two datasets are \textit{Blobs 1.3} and \textit{Blobs 1.7}, each of which has two dimensions and five classes, where the blobs are generated from a normal distribution with standard deviations of $1.3$ and $1.7$, respectively. The third and fourth are two class datasets named \textit{Spiral 2.5} and \textit{Spiral 3.5}, consisting of two interlocking spiral-shaped regions, each corresponding to one class, with the values $2.5$ and $3.5$ indicating the standard deviation from the center of the spiral, thus controlling the amount of overlap between the regions. These datasets are used for diagnostic purposes to understand the behaviour of the algorithm. Finally, we used the \emph{Blood Alcohol Concentration (BAC)} dataset, which is commonly utilized in decision-making and confidence estimation tasks. Data were first collected by Nugent and Cunningham \cite{nugent2005bloackbox} and can be used for regression and binary classification, depending on whether a threshold is set on the BAC level to distinguish between classes. Both the toy datasets, Blobs and Spirals, and BAC are made of $2000$ samples in total, $700$ are used for training, $300$ for validation and $1000$ in the test set.

For each dataset a feed forward neural network has been implemented, with a softmax function applied to the final logits. The MCP is extracted with a smooth maximum function, namely the \emph{Boltzman operator} \cite{asadi2017alternative}, to keep the confidence score differentiable with respect to the model parameters.
The Adam \cite{kingma2014adam} optimizer was used in the \emph{standard} training and in the inner loop of BO4SC (to optimize the model parameters $\theta$). All hyperparameters has been selected through a grid search. Besides the number of epochs, in the bilevel approach it is important to adjust the number of inner iterations ($T$) and the learning rate $\eta_{\mathbf{w}}$ for the update of the outer parameters.





\subsection{Confidence Estimation and Calibration}

What interests us in our experiments is assessing how well the methods predict calibrated confidence estimations.
We begin with a analysis using one of the toy datasets, where we can observe how well the models differentiate between high-confidence and low-confidence regions.

The toy dataset \emph{Blobs 1.7} provides an excellent case for this analysis. In Figure~\ref{fig:confidence_estimation_blobs1.7} below, we present an image made up of three different plots, each representing the confidence estimation results. These plots visually demonstrate the predicted confidence levels across the entire input space, highlighting areas where each model is more or less confident in its predictions.

\begin{figure}
    \centering
    \begin{subfigure}[b]{0.33\textwidth}
        \centering
        \includegraphics[width=\textwidth]{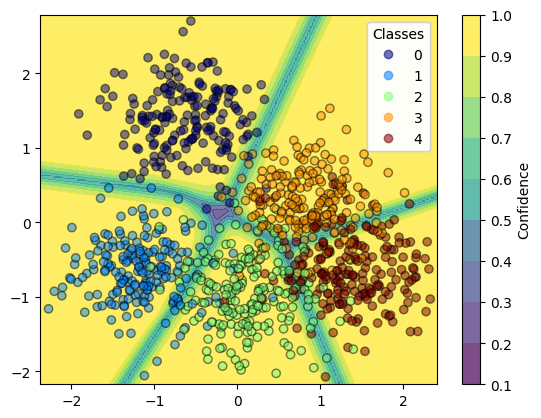}
        \caption{Standard}
        \label{fig:blobs17_simple}
    \end{subfigure}
    \hfill
    \begin{subfigure}[b]{0.33\textwidth}
        \centering
        \includegraphics[width=\textwidth]{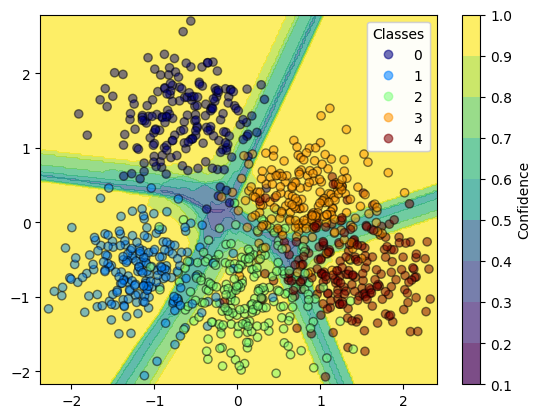}
        \caption{IsoReg}
        \label{fig:blobs17_bome}
    \end{subfigure}
    \hfill
    \begin{subfigure}[b]{0.33\textwidth}
        \centering
        \includegraphics[width=\textwidth]{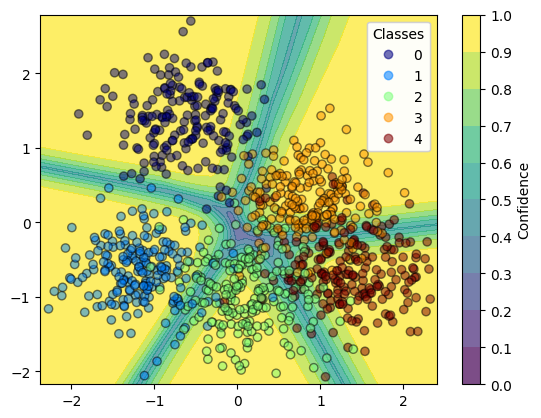}
        \caption{BO4SC}
        \label{fig:blobs17_explgrad}
    \end{subfigure}
    \caption{Confidence region estimation on the Blobs 1.7 dataset for differnent approaches. Each plot represents the spatial distribution of confidence levels across the dataset. The color in the background represents the confidence value that the model associates to a point that would be found in that place.}
    \label{fig:confidence_estimation_blobs1.7}
\end{figure}

The \textit{standard} model is highly confident in most regions, as indicated by the yellow areas. These regions reflect the areas where the model predicts class membership with high certainty (confidence value in $(0.9, 1]$). However, this confidence is sharply reduced in very narrow areas corresponding to the decision boundaries, represented by the green regions. These `lines' of uncertainty appear consistently thin across different parts of the dataset, irrespective of the degree of overlap between classes.
In contrast, the BO4SC model's confidence regions show a different pattern. Here, the uncertainty regions are considerably broader, especially in areas where the classes overlap more. This broader distribution of uncertainty better reflects the true complexity and intersections within the data, suggesting that the BO4SC approach is more sensitive to the nuances of the dataset's distribution. This ability, which also characterize the Isotonic Regression post-calibration method, represents a significant improvement over the Standard model, highlighting the advantages of a post- or self- calibration technique to address the confidence estimation challenge.

A more detailed examination using quantitative metrics is essential to rigorously evaluate the effectiveness of these methods and of bilevel optimization in producing well-calibrated models.

\begin{figure}
\centering
\begin{subfigure}[b]{0.33\textwidth} \centering
\includegraphics[width=\textwidth]{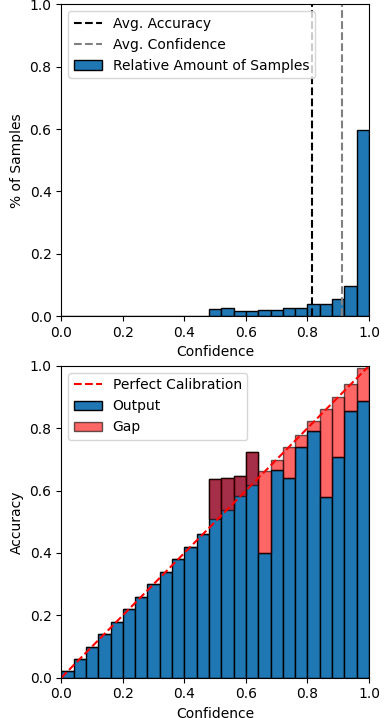}
\caption{Standard}
\end{subfigure}
\hfill
\begin{subfigure}[b]{0.33\textwidth}
\centering
\includegraphics[width=\textwidth]{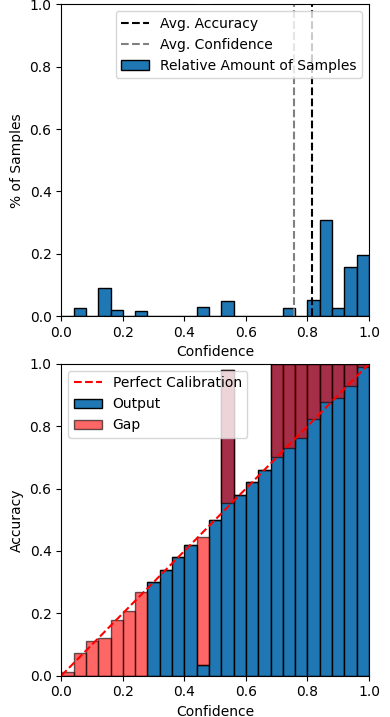}
\caption{IsoReg}
\end{subfigure}
\hfill
\begin{subfigure}[b]{0.33\textwidth}
\centering
\includegraphics[width=\textwidth]{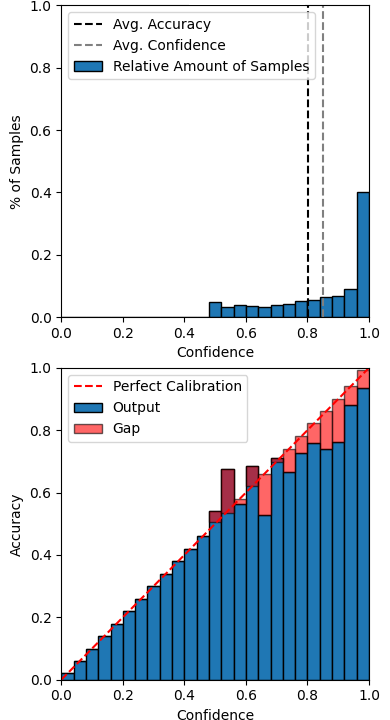}
\caption{BO4SC}
\end{subfigure}

\caption{Confidence Histograms (top) and Reliability Diagrams (bottom) for Spiral 3.5 test set.
Orange sections represent overconfident gap, while red represents underconfidence.
\label{fig:reliability_diagrams_spiral3.5}}
\end{figure}

The first step is to examine the confidence calibration through \emph{reliability diagrams} (Section \ref{calibration_metrics}) 
and \emph{confidence histograms}. These visual tools provide a direct representation of the relationship between predicted probabilities and actual observed frequencies, allowing for a straightforward assessment of a model's calibration. The plots are consistently similar in all datasets, and we present in Figure~\ref{fig:reliability_diagrams_spiral3.5} the reliability diagrams for the \textit{Spiral 3.5} dataset, which emphasize a drawback of Isotonic Regression.
A critical aspect to consider is the gap between the two dashed vertical lines in the confidence istogram: the darker line represents average accuracy, while the lighter grey line indicates the average confidence. For a model to be considered well-calibrated, these two lines should ideally overlap, or at least be very close to each other. The closer these lines are, the more aligned the model's predicted confidence is with its actual performance.
When we examine the toy datasets, the gap between these two lines becomes particularly noticeable. The Standard model consistently displays the largest gap between the average accuracy and the average confidence across all datasets. This wide gap, with the darker line staying on the left side, implies that the model's confidence scores are overly optimistic and do not accurately reflect its true performance. 

On the other hand, the BO4SC model shows the smallest gap, indicating that it has a more accurate alignment between confidence and accuracy. The IsoReg method also achieves a relatively close alignment between these two metrics. However, there is a nuanced difference between the confidence distribution obtained through bilevel optimization and the distribution achieved by post-calibration methods like IsoReg.
Although IsoReg effectively narrows the gap between accuracy and confidence, it does not always appropriately adjust confidence predictions. In the spiral dataset for example, the IsoReg model produces confidence scores that fall within the $(0, 0.5]$ range. Since these datasets are binary classification tasks, the minimum reasonable confidence score should be around $0.5$, reflecting the baseline probability of a random guess. The presence of lower confidence scores indicates an improper adjustment given by the IsoReg model, where it underestimates the confidence needed, thereby deviating from a reasonable calibration.

The reliability diagrams further reinforce the conclusions drawn from the confidence histograms. The Standard model demonstrates a clear tendency toward overconfidence. This is evident from the prevalence of orange gaps, especially in the higher confidence bins.
In contrast, the bilevel optimization approach exhibits much better calibration, with reliability diagrams visually more balanced.
Interestingly, while IsoReg effectively reduces the overconfidence seen in the Standard model, it introduces occasional calibration issues of its own. In particular, it may undercorrect or overcorrect certain confidence levels, leading to gaps that are not entirely aligned with the model's true accuracy.

With regard to confidence calibration metrics, we report in Table\ref{tab:comparison_metrics} the results for the \emph{Expected Calibration Error (ECE)} and the Accuracy of the models. The Expected Calibration Error (ECE) shows that the Bilevel Optimization method generally achieves lower values compared to the traditional Standard and IsoReg methods, indicating better calibration and more reliable confidence scores that are closer to the true probabilities, while keeping good accuracies overall.

\begin{table}
\caption{Comparison of Expected Calibration Error (ECE) and Accuracy across different datasets for Standard, IsoReg, and BO4SC methods. The best performance for each dataset and metric is highlighted in bold.}
\centering
{\footnotesize
\begin{tabular}{l|ccccc}
\toprule
\textbf{Method} & \textbf{Blobs 1.3} & \textbf{Blobs 1.7} & \textbf{Spiral 2.5} & \textbf{Spiral 3.5} & \textbf{BAC}  \\
\midrule
\multicolumn{6}{c}{\textbf{Expected Calibration Error (ECE)}} \\
\midrule
Standard     & 0.026 & 0.074 & 0.064 & 0.109 & 0.018  \\
IsoReg     & 0.023 & 0.039 & 0.039 & 0.143 & \bfseries 0.004  \\
BO4SC   & \bfseries 0.017 & \bfseries 0.016 & \bfseries 0.025 & \bfseries 0.067 & 0.012  \\
\midrule
\multicolumn{6}{c}{\textbf{Accuracy}} \\
\midrule
Standard     & \textbf{0.94}  & \textbf{0.876}  & 0.91   & \textbf{0.815}  & 0.989   \\
IsoReg    & \textbf{0.94}  & \textbf{0.876}  & 0.91   & \textbf{0.815}  & 0.989   \\
BO4SC  & 0.931  & 0.859  & \textbf{0.923} & 0.801  & \textbf{0.994}  \\
\bottomrule

\end{tabular}
}
\label{tab:comparison_metrics}
\end{table}

\subsection{Training Weights}

We can make some additional comments regarding the training approach that exploits bilevel optimization. One of these relates to the role of the \emph{weights} assigned to each training sample.
The weighted approach used allows the model to prioritize certain samples over others during training, potentially leading to better calibration and improved performance on more challenging or ambiguous classification.
By studying the evolution of these weights in BO4SC, we can better understand how the method operates.

\begin{figure}
\centering
\begin{subfigure}[b]{0.49\textwidth} \centering
\includegraphics[trim={0 0 0 24pt},clip,width=\textwidth]{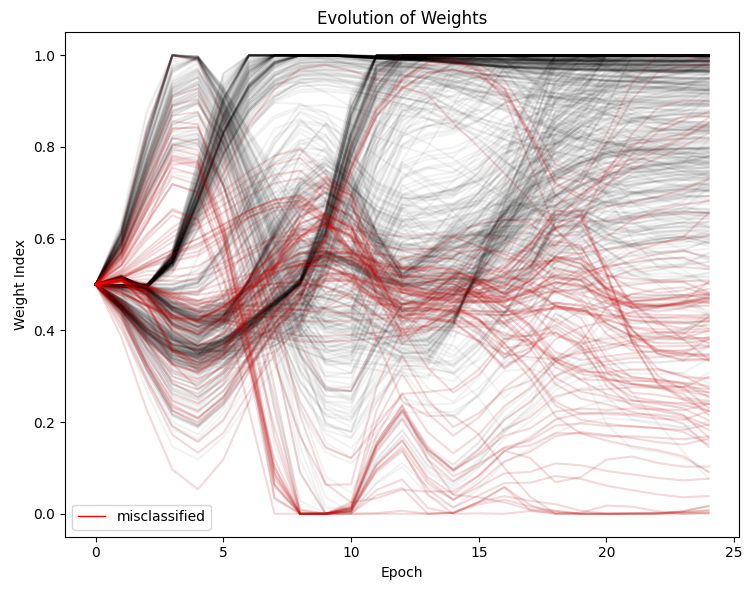}
\end{subfigure}
\hfill
\begin{subfigure}[b]{0.47\textwidth}
\centering
\includegraphics[trim={0 0 0 21pt},clip,width=\textwidth]{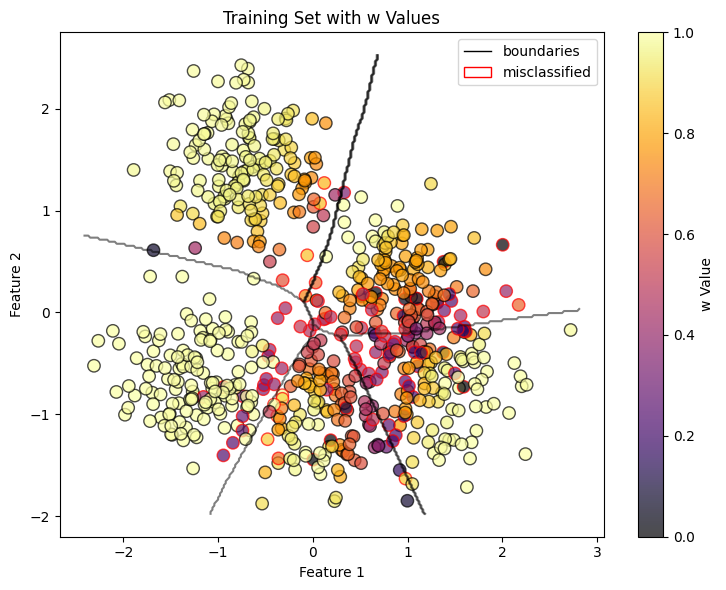}
\end{subfigure}

\caption{\textit{Left:} evolution of training \emph{weights} found by the BO4SC method for the Blobs 1.7 dataset (1 epoch unit = 10 training epochs). \textit{Right:} Final weight distribution.}
\label{fig:weights_evolutions}
\end{figure}

In Figure~\ref{fig:weights_evolutions} the history of the weights values (left panel) and their final distribution (right panel) are reported for the Blobs 1.7 dataset. In the left panel, the red lines indicate the weights associated with those samples that at the end result to be misclassified.
One can clearly see that the weights often move in groups, creating bundles of lines that follow the same trend. They might represent groups of samples close to each other that have the same characteristic or close in the variable space. The main observation is that most of the red lines end between $0$ and $0.5$, while the darker lines are mainly above the middle value.

Looking at the right panel of Figure~\ref{fig:weights_evolutions} one can observe that the BO4SC approach assigns a weight value of $1.0$ to samples that are clearly and confidently classified into a single class, typically those located near the center of each cluster, far from the decision boundaries.
As samples approach these boundaries, their weights decrease, converging towards 0.5 or even lower. This trend reflects BO4SC's strategy to diminish the influence of samples that are ambiguous or more likely to be misclassified.
This is visually evident, as many of these samples are marked with a red contour to indicate their misclassification, so belonging to a cluster of a different class, and appear as dark-colored (black) points, indicative of their low weight.

\section{Conclusions \label{sec:conclusions}}

In this article, we explored a novel bilevel optimization approach to address the challenge to self-calibrate a neural network in classification tasks. The objective was to improve the \emph{confidence} predicted by a model in such a way that it better reflects the actual accuracy and that it would be more meaningful in ambiguous scenarios. We made experimentation and analysis across a variety of datasets, ranging from toy datasets like Blobs and Spirals to more complex ones like BAC, and demonstrated the effectiveness of bilevel methods, particularly in their ability to refine confidence by dynamically adjusting sample weights during training.

We used the Expected Calibration Error (ECE) to quantitatively assess the models' performance. The consistent superiority of the bilevel approach over traditional methods highlights its ability to enhance classifier reliability while maintaining good accuracies overall.

The bilevel approach behaves well also when compared with post-calibration techniques. They present better results and, more importantly, they do not suffer from typical issues that show up when post-calibrating the confidence. In fact, we found that fine-tuning with post-calibration methods, like isotonic regression, occasionally leads to over-adjustments, resulting in overly cautious confidence estimates. For this reason, the confidence produced by the bilevel optimization methods would be more trustworthy in a real-world scenario.

While the results are promising, future research should focus on further refining these techniques. There is indeed still room for improvements on the computational side, i.e., executional time and memory performance of our bilevel approach are not always competitive with traditional training.
Another future research direction lies towards \emph{reject-option classification}, which allows models to refrain from making uncertain predictions.
\vspace{-0.7em}
\section*{Acknowledgements}
This work was supported by the STEM Challenge Fund 2023, University College Dublin.

\bibliography{sample-ceur}

\begin{thebibliography}{41}
\expandafter\ifx\csname natexlab\endcsname\relax\def\natexlab#1{#1}\fi
\providecommand{\url}[1]{\texttt{#1}}
\providecommand{\href}[2]{#2}
\providecommand{\path}[1]{#1}
\providecommand{\DOIprefix}{doi:}
\providecommand{\ArXivprefix}{arXiv:}
\providecommand{\URLprefix}{URL: }
\providecommand{\Pubmedprefix}{pmid:}
\providecommand{\doi}[1]{\href{http://dx.doi.org/#1}{\path{#1}}}
\providecommand{\Pubmed}[1]{\href{pmid:#1}{\path{#1}}}
\providecommand{\bibinfo}[2]{#2}
\ifx\xfnm\relax \def\xfnm[#1]{\unskip,\space#1}\fi
\bibitem[{Minderer et~al.(2021)Minderer, Djolonga, Romijnders, Hubis, Zhai,
  Houlsby, Tran, and Lucic}]{Minderer2021neurips}
\bibinfo{author}{M.~Minderer}, \bibinfo{author}{J.~Djolonga},
  \bibinfo{author}{R.~Romijnders}, \bibinfo{author}{F.~Hubis},
  \bibinfo{author}{X.~Zhai}, \bibinfo{author}{N.~Houlsby},
  \bibinfo{author}{D.~Tran}, \bibinfo{author}{M.~Lucic},
\newblock \bibinfo{title}{Revisiting the calibration of modern neural
  networks},
\newblock in: \bibinfo{booktitle}{Advances in Neural Information Processing
  Systems}, volume~\bibinfo{volume}{34}, \bibinfo{publisher}{Curran Associates,
  Inc.}, \bibinfo{year}{2021}, pp. \bibinfo{pages}{15682--15694}.
\bibitem[{Zhang et~al.(2023)Zhang, Xie, Li, Mei, and Liu}]{zhang2023survey}
\bibinfo{author}{X.-Y. Zhang}, \bibinfo{author}{G.-S. Xie},
  \bibinfo{author}{X.~Li}, \bibinfo{author}{T.~Mei}, \bibinfo{author}{C.-L.
  Liu},
\newblock \bibinfo{title}{A survey on learning to reject},
\newblock \bibinfo{journal}{Proceedings of the IEEE} \bibinfo{volume}{111}
  (\bibinfo{year}{2023}) \bibinfo{pages}{185--215}.
\bibitem[{Hendrickx et~al.(2024)Hendrickx, Perini, Van~der Plas, Meert, and
  Davis}]{hendrickx2024machine}
\bibinfo{author}{K.~Hendrickx}, \bibinfo{author}{L.~Perini},
  \bibinfo{author}{D.~Van~der Plas}, \bibinfo{author}{W.~Meert},
  \bibinfo{author}{J.~Davis},
\newblock \bibinfo{title}{Machine learning with a reject option: A survey},
\newblock \bibinfo{journal}{Machine Learning} \bibinfo{volume}{113}
  (\bibinfo{year}{2024}) \bibinfo{pages}{3073--3110}.
\bibitem[{Guo et~al.(2017)Guo, Pleiss, Sun, and
  Weinberger}]{guo2017calibration}
\bibinfo{author}{C.~Guo}, \bibinfo{author}{G.~Pleiss},
  \bibinfo{author}{Y.~Sun}, \bibinfo{author}{K.~Q. Weinberger},
\newblock \bibinfo{title}{On calibration of modern neural networks},
\newblock in: \bibinfo{booktitle}{International conference on machine
  learning}, \bibinfo{organization}{PMLR}, \bibinfo{year}{2017}, pp.
  \bibinfo{pages}{1321--1330}.
\bibitem[{Cosmides and Tooby(1996)}]{cosmides1996humans}
\bibinfo{author}{L.~Cosmides}, \bibinfo{author}{J.~Tooby},
\newblock \bibinfo{title}{Are humans good intuitive statisticians after all?
  rethinking some conclusions from the literature on judgment under
  uncertainty},
\newblock \bibinfo{journal}{cognition} \bibinfo{volume}{58}
  (\bibinfo{year}{1996}) \bibinfo{pages}{1--73}.
\bibitem[{Pedregosa(2016)}]{pedregosa2016hyperparameter}
\bibinfo{author}{F.~Pedregosa},
\newblock \bibinfo{title}{Hyperparameter optimization with approximate
  gradient},
\newblock in: \bibinfo{booktitle}{International conference on machine
  learning}, \bibinfo{organization}{PMLR}, \bibinfo{year}{2016}, pp.
  \bibinfo{pages}{737--746}.
\bibitem[{Franceschi et~al.(2017)Franceschi, Donini, Frasconi, and
  Pontil}]{franceschi2017forward}
\bibinfo{author}{L.~Franceschi}, \bibinfo{author}{M.~Donini},
  \bibinfo{author}{P.~Frasconi}, \bibinfo{author}{M.~Pontil},
\newblock \bibinfo{title}{Forward and reverse gradient-based hyperparameter
  optimization},
\newblock in: \bibinfo{booktitle}{International Conference on Machine
  Learning}, \bibinfo{organization}{PMLR}, \bibinfo{year}{2017}, pp.
  \bibinfo{pages}{1165--1173}.
\bibitem[{Franceschi et~al.(2018)Franceschi, Frasconi, Salzo, Grazzi, and
  Pontil}]{franceschi2018bilevel}
\bibinfo{author}{L.~Franceschi}, \bibinfo{author}{P.~Frasconi},
  \bibinfo{author}{S.~Salzo}, \bibinfo{author}{R.~Grazzi},
  \bibinfo{author}{M.~Pontil},
\newblock \bibinfo{title}{Bilevel programming for hyperparameter optimization
  and meta-learning},
\newblock in: \bibinfo{booktitle}{International conference on machine
  learning}, \bibinfo{organization}{PMLR}, \bibinfo{year}{2018}, pp.
  \bibinfo{pages}{1568--1577}.
\bibitem[{Jain and Shenoy(2022)}]{jain2022selective}
\bibinfo{author}{N.~Jain}, \bibinfo{author}{P.~Shenoy},
\newblock \bibinfo{title}{Selective classification using a robust meta-learning
  approach},
\newblock \bibinfo{journal}{arXiv preprint arXiv:2212.05987}
  (\bibinfo{year}{2022}).
\bibitem[{Weinberger and Saul(2009)}]{weinberger2009distance}
\bibinfo{author}{K.~Q. Weinberger}, \bibinfo{author}{L.~K. Saul},
\newblock \bibinfo{title}{Distance metric learning for large margin nearest
  neighbor classification.},
\newblock \bibinfo{journal}{Journal of machine learning research}
  \bibinfo{volume}{10} (\bibinfo{year}{2009}).
\bibitem[{Mendes~J{\'u}nior et~al.(2017)Mendes~J{\'u}nior, De~Souza, Werneck,
  Stein, Pazinato, De~Almeida, Penatti, Torres, and Rocha}]{mendes2017nearest}
\bibinfo{author}{P.~R. Mendes~J{\'u}nior}, \bibinfo{author}{R.~M. De~Souza},
  \bibinfo{author}{R.~d.~O. Werneck}, \bibinfo{author}{B.~V. Stein},
  \bibinfo{author}{D.~V. Pazinato}, \bibinfo{author}{W.~R. De~Almeida},
  \bibinfo{author}{O.~A. Penatti}, \bibinfo{author}{R.~d.~S. Torres},
  \bibinfo{author}{A.~Rocha},
\newblock \bibinfo{title}{Nearest neighbors distance ratio open-set
  classifier},
\newblock \bibinfo{journal}{Machine Learning} \bibinfo{volume}{106}
  (\bibinfo{year}{2017}) \bibinfo{pages}{359--386}.
\bibitem[{Jiang et~al.(2018)Jiang, Kim, Guan, and Gupta}]{jiang2018trust}
\bibinfo{author}{H.~Jiang}, \bibinfo{author}{B.~Kim},
  \bibinfo{author}{M.~Guan}, \bibinfo{author}{M.~Gupta},
\newblock \bibinfo{title}{To trust or not to trust a classifier},
\newblock \bibinfo{journal}{Advances in neural information processing systems}
  \bibinfo{volume}{31} (\bibinfo{year}{2018}).
\bibitem[{Mandelbaum and Weinshall(2017)}]{mandelbaum2017distance}
\bibinfo{author}{A.~Mandelbaum}, \bibinfo{author}{D.~Weinshall},
\newblock \bibinfo{title}{Distance-based confidence score for neural network
  classifiers},
\newblock \bibinfo{journal}{arXiv preprint arXiv:1709.09844}
  (\bibinfo{year}{2017}).
\bibitem[{Papernot and McDaniel(2018)}]{papernot2018deep}
\bibinfo{author}{N.~Papernot}, \bibinfo{author}{P.~McDaniel},
\newblock \bibinfo{title}{Deep k-nearest neighbors: Towards confident,
  interpretable and robust deep learning},
\newblock \bibinfo{journal}{arXiv preprint arXiv:1803.04765}
  (\bibinfo{year}{2018}).
\bibitem[{Gal and Ghahramani(2016)}]{gal2016dropout}
\bibinfo{author}{Y.~Gal}, \bibinfo{author}{Z.~Ghahramani},
\newblock \bibinfo{title}{Dropout as a bayesian approximation: Representing
  model uncertainty in deep learning},
\newblock in: \bibinfo{booktitle}{international conference on machine
  learning}, \bibinfo{organization}{PMLR}, \bibinfo{year}{2016}, pp.
  \bibinfo{pages}{1050--1059}.
\bibitem[{Blundell et~al.(2015)Blundell, Cornebise, Kavukcuoglu, and
  Wierstra}]{blundell2015weight}
\bibinfo{author}{C.~Blundell}, \bibinfo{author}{J.~Cornebise},
  \bibinfo{author}{K.~Kavukcuoglu}, \bibinfo{author}{D.~Wierstra},
\newblock \bibinfo{title}{Weight uncertainty in neural network},
\newblock in: \bibinfo{booktitle}{International conference on machine
  learning}, \bibinfo{organization}{PMLR}, \bibinfo{year}{2015}, pp.
  \bibinfo{pages}{1613--1622}.
\bibitem[{Kristiadi et~al.(2020)Kristiadi, Hein, and
  Hennig}]{kristiadi2020being}
\bibinfo{author}{A.~Kristiadi}, \bibinfo{author}{M.~Hein},
  \bibinfo{author}{P.~Hennig},
\newblock \bibinfo{title}{Being bayesian, even just a bit, fixes overconfidence
  in relu networks},
\newblock in: \bibinfo{booktitle}{International conference on machine
  learning}, \bibinfo{organization}{PMLR}, \bibinfo{year}{2020}, pp.
  \bibinfo{pages}{5436--5446}.
\bibitem[{Riquelme et~al.(2018)Riquelme, Tucker, and Snoek}]{riquelme2018deep}
\bibinfo{author}{C.~Riquelme}, \bibinfo{author}{G.~Tucker},
  \bibinfo{author}{J.~Snoek},
\newblock \bibinfo{title}{Deep bayesian bandits showdown: An empirical
  comparison of bayesian deep networks for thompson sampling},
\newblock \bibinfo{journal}{arXiv preprint arXiv:1802.09127}
  (\bibinfo{year}{2018}).
\bibitem[{Xia et~al.(2015)Xia, Cao, Wen, Hua, and Sun}]{xia2015learning}
\bibinfo{author}{Y.~Xia}, \bibinfo{author}{X.~Cao}, \bibinfo{author}{F.~Wen},
  \bibinfo{author}{G.~Hua}, \bibinfo{author}{J.~Sun},
\newblock \bibinfo{title}{Learning discriminative reconstructions for
  unsupervised outlier removal},
\newblock in: \bibinfo{booktitle}{Proceedings of the IEEE international
  conference on computer vision}, \bibinfo{year}{2015}, pp.
  \bibinfo{pages}{1511--1519}.
\bibitem[{Yoshihashi et~al.(2019)Yoshihashi, Shao, Kawakami, You, Iida, and
  Naemura}]{yoshihashi2019classification}
\bibinfo{author}{R.~Yoshihashi}, \bibinfo{author}{W.~Shao},
  \bibinfo{author}{R.~Kawakami}, \bibinfo{author}{S.~You},
  \bibinfo{author}{M.~Iida}, \bibinfo{author}{T.~Naemura},
\newblock \bibinfo{title}{Classification-reconstruction learning for open-set
  recognition},
\newblock in: \bibinfo{booktitle}{Proceedings of the IEEE/CVF Conference on
  Computer Vision and Pattern Recognition}, \bibinfo{year}{2019}, pp.
  \bibinfo{pages}{4016--4025}.
\bibitem[{Srivastava et~al.(2014)Srivastava, Hinton, Krizhevsky, Sutskever, and
  Salakhutdinov}]{srivastava2014dropout}
\bibinfo{author}{N.~Srivastava}, \bibinfo{author}{G.~Hinton},
  \bibinfo{author}{A.~Krizhevsky}, \bibinfo{author}{I.~Sutskever},
  \bibinfo{author}{R.~Salakhutdinov},
\newblock \bibinfo{title}{Dropout: a simple way to prevent neural networks from
  overfitting},
\newblock \bibinfo{journal}{The journal of machine learning research}
  \bibinfo{volume}{15} (\bibinfo{year}{2014}) \bibinfo{pages}{1929--1958}.
\bibitem[{Lakshminarayanan et~al.(2017)Lakshminarayanan, Pritzel, and
  Blundell}]{lakshminarayanan2017simple}
\bibinfo{author}{B.~Lakshminarayanan}, \bibinfo{author}{A.~Pritzel},
  \bibinfo{author}{C.~Blundell},
\newblock \bibinfo{title}{Simple and scalable predictive uncertainty estimation
  using deep ensembles},
\newblock \bibinfo{journal}{Advances in neural information processing systems}
  \bibinfo{volume}{30} (\bibinfo{year}{2017}).
\bibitem[{Bendale and Boult(2016)}]{bendale2016towards}
\bibinfo{author}{A.~Bendale}, \bibinfo{author}{T.~E. Boult},
\newblock \bibinfo{title}{Towards open set deep networks},
\newblock in: \bibinfo{booktitle}{Proceedings of the IEEE conference on
  computer vision and pattern recognition}, \bibinfo{year}{2016}, pp.
  \bibinfo{pages}{1563--1572}.
\bibitem[{De~Stefano et~al.(2000)De~Stefano, Sansone, and Vento}]{de2000reject}
\bibinfo{author}{C.~De~Stefano}, \bibinfo{author}{C.~Sansone},
  \bibinfo{author}{M.~Vento},
\newblock \bibinfo{title}{To reject or not to reject: that is the question-an
  answer in case of neural classifiers},
\newblock \bibinfo{journal}{IEEE Transactions on Systems, Man, and Cybernetics,
  Part C (Applications and Reviews)} \bibinfo{volume}{30}
  (\bibinfo{year}{2000}) \bibinfo{pages}{84--94}.
\bibitem[{DeGroot and Fienberg(1983)}]{degroot1983comparison}
\bibinfo{author}{M.~H. DeGroot}, \bibinfo{author}{S.~E. Fienberg},
\newblock \bibinfo{title}{The comparison and evaluation of forecasters},
\newblock \bibinfo{journal}{Journal of the Royal Statistical Society: Series D
  (The Statistician)} \bibinfo{volume}{32} (\bibinfo{year}{1983})
  \bibinfo{pages}{12--22}.
\bibitem[{Niculescu-Mizil and Caruana(2005)}]{niculescu2005predicting}
\bibinfo{author}{A.~Niculescu-Mizil}, \bibinfo{author}{R.~Caruana},
\newblock \bibinfo{title}{Predicting good probabilities with supervised
  learning},
\newblock in: \bibinfo{booktitle}{Proceedings of the 22nd international
  conference on Machine learning}, \bibinfo{year}{2005}, pp.
  \bibinfo{pages}{625--632}.
\bibitem[{Zadrozny and Elkan(2001)}]{zadrozny2001obtaining}
\bibinfo{author}{B.~Zadrozny}, \bibinfo{author}{C.~Elkan},
\newblock \bibinfo{title}{Obtaining calibrated probability estimates from
  decision trees and naive bayesian classifiers},
\newblock in: \bibinfo{booktitle}{Icml}, volume~\bibinfo{volume}{1},
  \bibinfo{year}{2001}, pp. \bibinfo{pages}{609--616}.
\bibitem[{Naeini et~al.(2015)Naeini, Cooper, and
  Hauskrecht}]{naeini2015obtaining}
\bibinfo{author}{M.~P. Naeini}, \bibinfo{author}{G.~Cooper},
  \bibinfo{author}{M.~Hauskrecht},
\newblock \bibinfo{title}{Obtaining well calibrated probabilities using
  bayesian binning},
\newblock in: \bibinfo{booktitle}{Proceedings of the AAAI conference on
  artificial intelligence}, volume~\bibinfo{volume}{29}, \bibinfo{year}{2015}.
\bibitem[{Platt et~al.(1999)}]{platt1999probabilistic}
\bibinfo{author}{J.~Platt}, et~al.,
\newblock \bibinfo{title}{Probabilistic outputs for support vector machines and
  comparisons to regularized likelihood methods},
\newblock \bibinfo{journal}{Advances in large margin classifiers}
  \bibinfo{volume}{10} (\bibinfo{year}{1999}) \bibinfo{pages}{61--74}.
\bibitem[{Kull et~al.(2017)Kull, Silva~Filho, and Flach}]{kull2017beta}
\bibinfo{author}{M.~Kull}, \bibinfo{author}{T.~Silva~Filho},
  \bibinfo{author}{P.~Flach},
\newblock \bibinfo{title}{Beta calibration: a well-founded and easily
  implemented improvement on logistic calibration for binary classifiers},
\newblock in: \bibinfo{booktitle}{Artificial intelligence and statistics},
  \bibinfo{organization}{PMLR}, \bibinfo{year}{2017}, pp.
  \bibinfo{pages}{623--631}.
\bibitem[{Wang et~al.(2019)Wang, Li, and Dang}]{wang2019calibrating}
\bibinfo{author}{Y.~Wang}, \bibinfo{author}{L.~Li}, \bibinfo{author}{C.~Dang},
\newblock \bibinfo{title}{Calibrating classification probabilities with
  shape-restricted polynomial regression},
\newblock \bibinfo{journal}{IEEE transactions on pattern analysis and machine
  intelligence} \bibinfo{volume}{41} (\bibinfo{year}{2019})
  \bibinfo{pages}{1813--1827}.
\bibitem[{Pan et~al.(2020)Pan, Ao, Tang, Lu, Liu, Xiao, and He}]{pan2020field}
\bibinfo{author}{F.~Pan}, \bibinfo{author}{X.~Ao}, \bibinfo{author}{P.~Tang},
  \bibinfo{author}{M.~Lu}, \bibinfo{author}{D.~Liu}, \bibinfo{author}{L.~Xiao},
  \bibinfo{author}{Q.~He},
\newblock \bibinfo{title}{Field-aware calibration: a simple and empirically
  strong method for reliable probabilistic predictions},
\newblock in: \bibinfo{booktitle}{Proceedings of The Web Conference 2020},
  \bibinfo{year}{2020}, pp. \bibinfo{pages}{729--739}.
\bibitem[{Zadrozny and Elkan(2002)}]{zadrozny2002transforming}
\bibinfo{author}{B.~Zadrozny}, \bibinfo{author}{C.~Elkan},
\newblock \bibinfo{title}{Transforming classifier scores into accurate
  multiclass probability estimates},
\newblock in: \bibinfo{booktitle}{Proceedings of the eighth ACM SIGKDD
  international conference on Knowledge discovery and data mining},
  \bibinfo{year}{2002}, pp. \bibinfo{pages}{694--699}.
\bibitem[{Kwon et~al.(2020)Kwon, Won, Kim, and Paik}]{kwon2020uncertainty}
\bibinfo{author}{Y.~Kwon}, \bibinfo{author}{J.-H. Won}, \bibinfo{author}{B.~J.
  Kim}, \bibinfo{author}{M.~C. Paik},
\newblock \bibinfo{title}{Uncertainty quantification using bayesian neural
  networks in classification: Application to biomedical image segmentation},
\newblock \bibinfo{journal}{Computational Statistics \& Data Analysis}
  \bibinfo{volume}{142} (\bibinfo{year}{2020}) \bibinfo{pages}{106816}.
\bibitem[{Domke(2012)}]{domke2012generic}
\bibinfo{author}{J.~Domke},
\newblock \bibinfo{title}{Generic methods for optimization-based modeling},
\newblock in: \bibinfo{booktitle}{Artificial Intelligence and Statistics},
  \bibinfo{organization}{PMLR}, \bibinfo{year}{2012}, pp.
  \bibinfo{pages}{318--326}.
\bibitem[{Maclaurin et~al.(2015)Maclaurin, Duvenaud, and
  Adams}]{maclaurin2015gradient}
\bibinfo{author}{D.~Maclaurin}, \bibinfo{author}{D.~Duvenaud},
  \bibinfo{author}{R.~Adams},
\newblock \bibinfo{title}{Gradient-based hyperparameter optimization through
  reversible learning},
\newblock in: \bibinfo{booktitle}{International conference on machine
  learning}, \bibinfo{organization}{PMLR}, \bibinfo{year}{2015}, pp.
  \bibinfo{pages}{2113--2122}.
\bibitem[{Ren* et~al.(2023)Ren*, Feng*, Liu*, Pan*, Fu, Mai, and
  Yang}]{JMLR:TorchOpt}
\bibinfo{author}{J.~Ren*}, \bibinfo{author}{X.~Feng*},
  \bibinfo{author}{B.~Liu*}, \bibinfo{author}{X.~Pan*},
  \bibinfo{author}{Y.~Fu}, \bibinfo{author}{L.~Mai}, \bibinfo{author}{Y.~Yang},
\newblock \bibinfo{title}{Torchopt: An efficient library for differentiable
  optimization},
\newblock \bibinfo{journal}{Journal of Machine Learning Research}
  \bibinfo{volume}{24} (\bibinfo{year}{2023}) \bibinfo{pages}{1--14}.
\bibitem[{Paszke et~al.(2017)Paszke, Gross, Chintala, Chanan, Yang, DeVito,
  Lin, Desmaison, Antiga, and Lerer}]{paszke2017automatic}
\bibinfo{author}{A.~Paszke}, \bibinfo{author}{S.~Gross},
  \bibinfo{author}{S.~Chintala}, \bibinfo{author}{G.~Chanan},
  \bibinfo{author}{E.~Yang}, \bibinfo{author}{Z.~DeVito},
  \bibinfo{author}{Z.~Lin}, \bibinfo{author}{A.~Desmaison},
  \bibinfo{author}{L.~Antiga}, \bibinfo{author}{A.~Lerer},
\newblock \bibinfo{title}{Automatic differentiation in pytorch},
\newblock in: \bibinfo{booktitle}{NIPS-W}, \bibinfo{year}{2017}.
\bibitem[{Nugent and Cunningham(2005)}]{nugent2005bloackbox}
\bibinfo{author}{C.~Nugent}, \bibinfo{author}{P.~Cunningham},
\newblock \bibinfo{title}{A case-based explanation system for black-box
  systems},
\newblock \bibinfo{journal}{Artif. Intell. Rev.} \bibinfo{volume}{24}
  (\bibinfo{year}{2005}) \bibinfo{pages}{163--178}.
\bibitem[{Asadi and Littman(2017)}]{asadi2017alternative}
\bibinfo{author}{K.~Asadi}, \bibinfo{author}{M.~L. Littman},
\newblock \bibinfo{title}{An alternative softmax operator for reinforcement
  learning},
\newblock in: \bibinfo{booktitle}{International Conference on Machine
  Learning}, \bibinfo{organization}{PMLR}, \bibinfo{year}{2017}, pp.
  \bibinfo{pages}{243--252}.
\bibitem[{Kingma and Ba(2017)}]{kingma2014adam}
\bibinfo{author}{D.~P. Kingma}, \bibinfo{author}{J.~Ba},
\newblock \bibinfo{title}{Adam: A method for stochastic optimization},
\newblock \bibinfo{journal}{arXiv preprint arXiv:1412.6980}
  (\bibinfo{year}{2017}).

\end{thebibliography}




\end{document}